\documentclass[10pt, a4paper]{article}

\usepackage{amsmath}
\usepackage{tikz}
\usepackage{booktabs}
\usepackage{tabularx}
\usepackage{siunitx}
\usepackage{dblfloatfix}
\usetikzlibrary{arrows.meta,positioning,fit,shapes,calc}

\usepackage[final]{lrec2026}

\title{Beyond Catalogue Counts: the Dataset Visibility Asymmetry in Low-Resource Multilingual NLP}

\name{Zhiyin Tan$^1$, Changxu Duan$^2$} 

\address{$^1$L3S Research Center, Leibniz University Hannover, Appelstraße 9A, 30167 Hannover, Germany \\
         zhiyin.tan@l3s.de \\
         $^2$Technische Universität Darmstadt, Residenzschloss 1, 64283 Darmstadt, Germany \\
         duan@linglit.tu-darmstadt.de}

\abstract{
Multilingual NLP often relies on dataset counts from centralized catalogues to characterize which languages are resource-rich or resource-poor. 
However, these catalogues record only one layer of dataset visibility: what has been registered or institutionally distributed. 
They do not necessarily reflect which datasets are created, cited, or reused in the research literature.
To examine this gap, we combine a catalogue-based baseline with literature-backed evidence of dataset circulation. 
We introduce the Resource Density Index (RDI), defined as the number of catalogued datasets per one million speakers, and compute it for the 200 most widely spoken languages in \textit{Ethnologue}. 
Among them, 118 languages (59\%) have an average RDI of zero across the LRE Map and the Linguistic Data Consortium (LDC), and another 23 fall below 0.1, corresponding to at most one catalogued dataset per ten million speakers.
We then apply an LLM-assisted citation-mining pipeline over the Semantic Scholar corpus to these 141 low-visibility languages. 
After manual validation and consolidation, we identify 609 unique datasets across 53 languages, of which 356 remain openly accessible through working public links. 
These results reveal a substantial visibility gap: many large-speaker languages appear data-poor in catalogue records yet show clear evidence of dataset activity in the research literature. 
Our findings suggest that multilingual data scarcity should be understood not only as a production problem, but also as a question of documentation, discoverability, and long-term accessibility.
Code and data are publicly available at \url{https://github.com/zhiyintan/dataset-visibility-asymmetry}.
\\ \newline  }

\begin{document}

\maketitleabstract

\section{Introduction}
\label{sec:intro}

Linguistic datasets are a central part of multilingual natural language processing (NLP). 
They shape which languages can be modeled, benchmarked, and evaluated, and therefore influence which languages are most visible in deployed language technologies \citep{PAULLADA2021100336,blasi-etal-2022-systematic}. 
Over the past decade, the field has invested heavily in documenting and cataloguing language resources so that datasets can be discovered, cited, and reused. 
This goal is explicit in infrastructures such as the LRE Map \citep{calzolari-etal-2010-lrec,del-gratta-etal-2018-lremap,DelGratta2021LREMap}, CLARIN,\footnote{\url{https://www.clarin.eu/}} and institutional repositories such as the Linguistic Data Consortium (LDC).\footnote{\url{https://catalog.ldc.upenn.edu/}} 
These infrastructures have become important reference points in dataset audits and multilingual surveys that discuss which languages appear well documented and which appear under-resourced \citep[e.g.,][]{joshi2020state,hedderich-etal-2021-survey,ranathunga-de-silva-2022-languages}. 

At the same time, catalogue counts should not be mistaken for a complete picture of multilingual resource availability. 
Catalogues record a particular layer of visibility: what has been submitted, curated, or institutionally distributed. 
They do not necessarily capture all datasets that are created, cited, reused, or circulated in research practice. 
Coverage may depend on voluntary submission, venue participation, licensing arrangements, repository scope, and the persistence of metadata over time. 
As a result, catalogue statistics are highly informative, but they remain an incomplete proxy for the broader landscape of dataset documentation and research circulation.

This distinction matters because the term \textit{low-resource} is itself broader than any single count. 
A recent qualitative survey of 150 papers shows that definitions of low-resourcedness draw on at least four dimensions: socio-political conditions, the availability of human and digital resources, artifacts such as datasets and tools, and community agency \citep{nigatu-etal-2024-zenos}. 
Within NLP practice, however, datasets occupy a particularly influential position. 
They are one of the main artifacts through which languages become benchmarked, compared, and discussed. 
For that reason, dataset counts often function as a practical signal of whether a language appears resource-rich or resource-poor, even though low-resourcedness cannot be reduced to dataset count alone.

The difficulty is that dataset visibility depends not only on whether a dataset exists, but also on whether it is recorded in the infrastructures that researchers typically consult. 
For instance, the LRE Map lists 31 Indonesian resources, the community-driven NusaCrowd \citep{cahyawijaya-etal-2023-nusacrowd} reports 137, while our literature-based audit identifies 196 distinct Indonesian datasets with validated evidence of use in research. 
For Marathi, the corresponding numbers are 14 in the catalogue view and 41 in our validated inventory. 
These examples do not show that catalogues are wrong, nor that all resource counts should collapse into a single harmonized total. 
They show something more specific: different infrastructures surface different layers of evidence, and catalogue-side visibility can fall far below what becomes visible through systematic tracing of the research literature.

This gap has practical consequences. 
Labels such as \textit{low-resource} shape how shared tasks are framed \citep{mia-2022-multilingual}, which languages are prioritized for benchmark construction and model evaluation, and how multilingual inequality is described in NLP \citep{blasi-etal-2022-systematic,yu-etal-2022-beyond}. 
If the evidence base behind those judgments is incomplete, then downstream claims about representation, resource scarcity, and language inequality may also be incomplete.

In this paper, we focus on one specific part of that broader problem: the visibility of datasets as documented and used artifacts in multilingual NLP. 
We ask two related questions. 
First, what picture of resource availability emerges from major catalogues when dataset counts are normalized by speaker population? 
Second, how does that picture change when we trace datasets that are visible through citation evidence in the research literature?

To answer these questions, we introduce a population-normalized metric, the \textbf{Resource Density Index (RDI)}, which measures the number of catalogued datasets per one million speakers across the 200 most widely spoken languages in the 2025 \textit{Ethnologue} list. 
We then complement this catalogue-based view with a citation-based audit of the Semantic Scholar corpus, adapted from our prior work on research-question-driven dataset discovery \citep{duan-tan-2025-soft,tan-duan-2025-citation}. 
Rather than estimating the full universe of existing resources, this second component identifies a narrower but verifiable layer of evidence: datasets cited, described, or reused in the literature and manually validated as language-specific resources. 
Together, these two perspectives allow us to compare catalogue-documented visibility with literature-backed evidence of dataset circulation.
This work makes three contributions:

\begin{enumerate}
    \item \textbf{A population-normalized view of catalogue visibility.}
    We introduce the Resource Density Index (RDI), a transparent metric for comparing catalogue-documented datasets across the 200 most widely spoken languages. 
    By combining demographic information from \textit{Ethnologue} with entries from the LRE Map and the LDC, RDI makes it possible to compare documentation density across languages of very different scale and to identify the segment of the distribution that appears nearly invisible in major catalogues.

    \item \textbf{A citation-validated audit of dataset circulation.}
    We adapt a citation-based dataset discovery framework to construct a language-by-language inventory of datasets evidenced in the research literature. 
    The resulting inventory is manually validated, deduplicated, and enriched with accessibility metadata, providing a usage-centered complement to catalogue-based documentation.

    \item \textbf{Evidence for visibility and accessibility gaps.}
    By comparing catalogue-based RDI estimates with citation-grounded evidence of datasets cited, described, or reused in the literature, we show that catalogue visibility and research circulation often diverge: many languages with zero or near-zero catalogue presence nonetheless have datasets documented and used in the research literature. 
    This finding suggests that part of the scarcity inferred from catalogue-side evidence may instead reflect gaps in documentation, indexing, and preservation that obscure resources already in research circulation. 
    We further distinguish between documented dataset existence and present-day open accessibility, highlighting how multilingual dataset inequality is shaped not only by dataset creation, but also by documentation, discoverability, and long-term access.
    
\end{enumerate}

Taken together, these contributions provide a demographically grounded and empirically validated framework for reassessing multilingual dataset visibility. 
They do not redefine low-resourcedness in general. 
Rather, they clarify one important part of it: how the field sees, counts, and compares datasets across languages.

\section{Related Work}
\label{sec:related}

\paragraph{Catalogue infrastructures and resource visibility.}
Studies of multilingual resource availability often rely on large cataloguing infrastructures that document and index language resources. 
Systematic documentation has long been a central goal of the language resources and evaluation community. 
The LRE Map \citep{calzolari-etal-2010-lrec,del-gratta-etal-2018-lremap,DelGratta2021LREMap}, developed under ELRA, was designed as a community-facing registry of corpora, tools, and lexica linked to research publications, with the aim of improving discoverability, citation, and reproducibility. 
Institutional repositories such as the Linguistic Data Consortium (LDC) and infrastructures such as CLARIN maintain curated catalogues with standardized metadata, persistent identifiers, and formal distribution procedures. \footnote{\url{https://www.clarin.eu/content/component-metadata}}
Related initiatives, including OLAC metadata standards and the Component Metadata Infrastructure, further reflect a broader effort to support interoperable resource description and long-term indexing.\footnote{\url{http://www.language-archives.org/OLAC/metadata.html}}
Beyond their role as storage or distribution platforms, these infrastructures function as a key layer of \emph{resource visibility}. 
Catalogue entries determine which datasets are easily discoverable, citable, and legible to subsequent surveys of multilingual NLP. 
For this reason, catalogue statistics are frequently used when researchers characterize languages as low-resource or under-resourced \citep[e.g.,][]{joshi2020state,hedderich-etal-2021-survey,ranathunga-de-silva-2022-languages}. 
At the same time, catalogue coverage is necessarily partial. 
Which resources appear in a given catalogue is shaped by institutional curation criteria, the venues through which submissions are solicited, and the distribution models each repository supports. 
Catalogue records therefore provide a valuable but incomplete view of multilingual dataset availability: they reflect what has been registered within particular infrastructures, not the full set of datasets that may exist or circulate in research practice.

\paragraph{Community and regional curation.}
Alongside these infrastructures, several initiatives have sought to document language resources within specific linguistic regions or research communities. 
For African languages, \citet{emezue2020lanfrica} introduce \textsc{Lanfrica}, a knowledge graph that aggregates NLP datasets, tools, and benchmarks produced by African research groups. 
Community collaborations such as Masakhane and HausaNLP curate task-specific corpora and resources for languages of the African continent \citep{muhammad-etal-2025-hausanlp}. 
In Southeast Asia, \citet{cahyawijaya-etal-2023-nusacrowd} present NusaCrowd, a large-scale effort to collect datasets for Indonesian and related regional languages.
These initiatives illustrate how dataset documentation can also emerge through decentralized curation practices. 
However, such efforts typically focus on particular regions or language families and therefore do not provide a global view of how catalogue records relate to the datasets actually used in research.

\paragraph{Defining and measuring low-resource.}
A parallel line of work examines how to define and measure linguistic resource inequality. 
Early work often equated low-resource status with the scarcity of labeled or parallel data, typically framing the problem at the level of specific tasks such as tagging, parsing, or machine translation \citep{hedderich-etal-2021-survey}. 
\citet{joshi2020state} highlighted the strong concentration of NLP research output in a small set of languages, particularly English. 
Subsequent studies have shown that disparities in language technology correlate with broader economic and infrastructural factors rather than speaker population alone \citep{blasi-etal-2022-systematic}. 
Other work has examined how resource coverage relates to demographic and socioeconomic indicators, demonstrating that languages with large speaker populations are not necessarily well represented in available datasets or benchmarks \citep{ranathunga-de-silva-2022-languages}. 
More recently, qualitative analyses of the literature have emphasized that low-resourcedness is a multidimensional concept. 
For example, \citet{nigatu-etal-2024-zenos} survey 150 papers and show that definitions of low-resourcedness draw on a range of factors, including socio-political context, human and digital infrastructure, available language technology artifacts, and community agency. 
Complementing this perspective, \citet{yu-etal-2022-beyond} analyze multilingual dataset construction practices and show that resource inequality is not only quantitative but also qualitative: datasets for many languages differ in provenance, annotation processes, and intended usage. 
Together, these studies highlight the complexity of defining “low-resource.” 
At the same time, in practical NLP discourse, dataset availability remains one of the most commonly used operational signals when describing whether a language appears resource-rich or resource-poor.

\paragraph{Summary.}
Taken together, prior work has established both the infrastructures that document language resources and the conceptual frameworks used to describe linguistic resource inequality. 
Yet an important question remains largely unexamined: how closely do catalogue-based representations of multilingual resources correspond to the datasets that actually circulate in research practice? 
If catalogue records function as a primary layer of resource visibility, discrepancies between catalogue coverage and research usage can reshape how languages are perceived within the field. 
Our study addresses this question by combining population-normalized catalogue analysis with citation-based evidence of dataset usage, enabling a systematic examination of where documented visibility diverges from empirical research activity.

\section{Methodology}
\label{sec:method}

Our goal is to characterize multilingual dataset visibility in a way that is comparable across large language communities, grounded in documented evidence, and sensitive to the gap between catalogue records and research circulation.
To do so, we combine two complementary views. 
First, we construct a population-normalized baseline from two major catalogue infrastructures. 
Second, we trace datasets that are visible through citation evidence in the research literature and manually validate the resulting inventory. 
This section introduces the population-normalized metric (Section~\ref{subsec:rdi}), the catalogue baseline construction (Section~\ref{subsec:catalog-baseline}), the citation-based dataset discovery procedure (Section~\ref{subsec:citation-mining}), the manual validation and consolidation protocol (Section~\ref{subsec:validation}), and the dataset attribute annotation procedure (Section~\ref{subsec:annotation}).

\subsection{Resource Density Index (RDI)}
\label{subsec:rdi}

To compare documented dataset availability across languages of very different scale, we define the \textit{Resource Density Index} (RDI) for a language $i$ as

\[
\text{RDI}_i = \frac{\text{datasets}_i}{\text{population}_i / 10^6},
\]
that is, the number of documented datasets per one million speakers.
Population is used here as a normalization factor for comparability, not as a claim about what any language deserves, and not as an assumption that larger speaker communities automatically yield richer dataset ecosystems. 
Rather, prior work on low-resourcedness shows that resource status is often discussed partly in relation to the scale of human and digital resources available to a language community \citep{nigatu-etal-2024-zenos}. 
Speaker population therefore provides a useful background variable against which catalogue documentation can be compared across languages of very different size.

Population estimates are taken from the \textit{Ethnologue} list of the 200 most widely spoken languages worldwide~\footnote{\url{https://www.ethnologue.com/insights/ethnologue200/}} (2025 edition).
This gives us a fixed comparison universe that is globally meaningful yet still feasible for careful manual analysis. 
For each of these 200 languages, we count the number of datasets reported in two widely used multilingual resource catalogues: the LRE Map \citep{calzolari-etal-2010-lrec,del-gratta-etal-2018-lremap,DelGratta2021LREMap} and the Linguistic Data Consortium (LDC).
We compute an RDI separately for each catalogue and average the two values to obtain a single catalogue-based estimate per language. 
Because large-scale cross-repository deduplication is unreliable in the absence of standardized resource identifiers and aligned metadata, we treat the two catalogues as complementary observations of documented language visibility. 
The averaged value should therefore be interpreted as a conservative visibility baseline rather than an estimate of the true union of resources across repositories.

Across the 200 languages in this comparison set, 118 have an average RDI of exactly 0.0, meaning that neither the LRE Map nor the LDC lists any dataset for them. 
A further 23 languages fall in the range $(0, 0.1)$. 
An RDI of 0.1 corresponds to approximately one catalogued dataset per ten million speakers. 
In subsequent analyses, we focus on languages whose average catalogue-based RDI falls below 0.1. 
This threshold does not imply that languages above it are well-resourced. 
Rather, it identifies the segment of the distribution that appears nearly invisible in major catalogues and is therefore most likely to be affected by documentation gaps.

\subsection{Catalogue Baseline and Language Normalization}
\label{subsec:catalog-baseline}

The catalogue baseline is derived from two sources with different institutional logics of visibility. 
The LRE Map is a community-driven registry populated largely through author submissions associated with venues such as LREC. 
The LDC, by contrast, is a curated institutional repository that distributes resources under managed licensing and access conditions. 
We use both because they capture different forms of documented presence: reported registration in one case, and maintained institutional distribution in the other.

Because language names in these catalogues are not always standardized, we manually normalize them to align with \textit{Ethnologue}'s language inventory.
This normalization collapses spelling variants and macro-labels where the catalogue clearly refers to the same language variety.
For example, \textit{Modern Greek} and \textit{Greek} are merged under \textit{Greek}; \textit{Persian} and \textit{Iranian Persian} are merged under \textit{Iranian Persian}; \textit{Brazilian Portuguese} is merged into \textit{Portuguese}; and \textit{Uighur} is merged into \textit{Uyghur}. 
For Chinese, catalogue entries labeled \textit{Chinese}, \textit{Mandarin}, and \textit{Mandarin Chinese} are merged as \textit{Mandarin} for comparability with \textit{Ethnologue}. 
For Arabic, \textit{Levantine Arabic}, \textit{North Levantine Arabic}, and \textit{South Levantine Arabic} are merged into \textit{Levantine Arabic}. 
Capitalization differences such as \textit{french} versus \textit{French} and \textit{italian} versus \textit{Italian} are resolved to a single canonical label.

When a catalogue uses an underspecified umbrella label that conflates sociolinguistically distinct varieties, however, we do not force a more specific assignment. 
For example, the LRE Map contains entries labeled \textit{Punjabi} and \textit{Pashto} without consistently distinguishing varieties that differ in standardization and script. 
In such cases, if the available metadata do not support a reliable mapping to a specific \textit{Ethnologue} variety, we leave the entry at the broader level rather than guessing. 
This conservative policy avoids inflating per-language counts through overassignment. 
The same principle applies to other cases in which catalogue labels underspecify dialectal, pluricentric distinctions.

The resulting normalized mapping gives us, for each of the 200 languages in our comparison set, a dataset count from the LRE Map and a dataset count from the LDC. 
These counts form the basis of the catalogue-side RDI values defined above.

\subsection{Citation-based Dataset Discovery}
\label{subsec:citation-mining}

Catalogue counts tell us what has been recorded in major infrastructures. 
They do not tell us, by themselves, which datasets have entered research circulation. 
To address this gap, we build on a citation-based dataset discovery framework introduced in our prior work \citep{duan-tan-2025-soft,tan-duan-2025-citation}. 
That earlier framework was designed for research-question-driven dataset discovery. 
Here, we adapt it to a different task: constructing a language-by-language inventory of datasets that are visible through citation evidence in the research literature.

Although the pipeline is general, the present analysis focuses on the 141 languages whose average catalogue-based RDI falls below 0.1. 
This design choice reflects an asymmetry in interpretive impact. 
For languages that are already well documented in catalogues, an additional missing dataset is usually incremental. 
For languages with zero or near-zero catalogue visibility, by contrast, even a small number of overlooked datasets can substantially change how the resource landscape is perceived. 
Citation-based discovery is therefore most informative in precisely the low-visibility segment where catalogue-side evidence is weakest.

Following the M3D framework (Multi-Disciplinary Dataset Discovery)\footnote{\url{https://github.com/Fireblossom/citation-context-dataset-discovery}} proposed by \citet{tan-duan-2025-citation}, we retrieve candidate papers from Semantic Scholar \citep{kinney2023semantic} for each of these 141 languages in which the language name appears together with resource-related terms such as \textit{corpus}, \textit{dataset}, or \textit{data}.
To keep the procedure computationally feasible and focused on the most visible literature, we apply a top-$k$ filter and retain at most the top $k=400$ papers per language. 
From this candidate pool, we extract both outgoing references (papers claiming to use or build on a dataset) and incoming references (papers citing a dataset-creation paper).
For each citation context, we use Qwen2.5-72B in a zero-shot classification setup to decide whether the cited item is a language-specific dataset rather than some other kind of scholarly artifact, such as a model, a toolkit, an evaluation metric, or a general software library. 
The full technical details of the prompting scheme, schema-guided extraction, and provenance-preserving entity resolution procedure are described in \citet{tan-duan-2025-citation}. 
In the present work, however, automatic classification is only the first stage. 

This procedure yields 812 candidate dataset mentions across the 141 target languages. 
Each candidate consists of a tuple of the form \{language, citing paper, cited paper, citation context\}. 
These candidates are then manually validated, filtered, and consolidated as described in Section~\ref{subsec:validation}. 
The result is not an estimate of the full universe of multilingual datasets. 
Rather, it is a bounded and manually verified inventory of datasets that leave recoverable traces in the research literature within our search scope.

\subsection{Manual Validation and Dataset Consolidation}
\label{subsec:validation}

To ensure the quality of the citation-based inventory, we apply a multi-stage manual validation procedure to all 812 candidate dataset mentions produced by the automatic discovery stage. 
The purpose of this procedure is threefold: to verify that the candidate refers to a genuine dataset, to remove false positives and unresolvable cases, and to consolidate multiple mentions of the same underlying resource.

\noindent\textbf{Step 1:} The pipeline produces 812 candidate dataset mentions. 

\noindent\textbf{Step 2:} Each candidate is then manually inspected by annotators using the citation context together with the relevant citing and cited paper metadata. In this stage, we remove 101 unconfirmable mentions whose contexts do not provide sufficient evidence to determine whether a dataset is being referenced, and 44 non-dataset resources such as books, dictionaries, software toolkits, and general-purpose libraries. After this first filtering step, 667 candidate mentions are confirmed as genuine dataset mentions. We define the precision of the automatic mention-classification stage as the ratio of genuine mentions to initial candidates, yielding $667 / 812 = 82.14\%$.

\noindent\textbf{Step 3:} The 667 genuine mentions still contain multiple references to the same underlying resource. We therefore consolidate duplicate surface forms and closely overlapping references when they clearly point to the same dataset. For example, \textit{The Penn Treebank} and \textit{PTB} are merged if they share the same underlying Semantic Scholar evidence and citation context. We also exclude direct translations or simple reslicings of existing datasets when they do not constitute distinct dataset projects in their own right. This consolidation step removes 58 duplicate or non-distinct entries.

\noindent\textbf{Step 4:} After filtering and consolidation, the final inventory contains 609 unique datasets across 53 of the 141 low-visibility languages investigated. This inventory forms the basis of all subsequent analyses.
    
We emphasize precision rather than recall. 
Estimating recall would require exhaustive annotation over all potentially relevant multilingual papers, which is prohibitively expensive at this scale. 
Our objective is therefore not to build a complete dataset-mining system, but to construct a high-confidence inventory of datasets that are visible in research circulation, including many that do not appear in centralized catalogues.

\subsection{Dataset Attribute Annotation}
\label{subsec:annotation}

For each of the 609 validated datasets, we record two additional attributes used in subsequent analyses, e.g.\ a temporal attribute (canonical source paper and emergence year) and an accessibility attribute (whether the dataset is openly obtainable at the time of annotation). 
These two dimensions are independent and are assessed separately for every dataset in the inventory.

\paragraph{Temporal attribution.}
For each dataset, we attempt to identify a unique canonical source paper whose publication year serves as the dataset's emergence year.
Of the 609 datasets, 549 can be linked to such a paper. 
The remaining 60 either correspond to multiple plausible source papers or are evidenced primarily through project pages or repositories without a single definitive publication.

\paragraph{Accessibility.}
Independently, we perform manual URL verification for each of the 609 datasets. 
An annotator follows associated links and checks whether the link resolves, whether the destination page refers to the same dataset, and whether the page provides either a direct download path or an access procedure that does not require institutional affiliation or paywalled licensing. 
This notion of open availability is intentionally narrower than institutional availability under managed licensing. 
For example, a dataset distributed through the LDC may remain obtainable under a paid access model while still counting as not openly available under our criterion.
We classify a dataset as \textit{openly available} if at least one working URL resolves and exposes either a direct download or an unrestricted access procedure. 
We classify a dataset as \textit{not openly available} if the dataset is clearly evidenced in the literature but no working link can be found, the link is dead, or access depends on closed distribution arrangements.
Applying this scheme, 356 of the 609 datasets have a working public download link at the time of annotation. 
The remaining 253 are documented in the literature but are either no longer retrievable online or available only through gated channels. 
This distinction allows us to separate literature-backed existence from present-day public accessibility.

Finally, the resulting citation-based inventory provides evidence of downstream research circulation. 
Overall, the datasets we identify are cited, described, or reused in the Semantic Scholar papers. 
This gives us an empirical view of which datasets are visible in research practice, regardless of whether they appear in centralized catalogue records.
The complete per-language dataset lists and all manual annotation details are publicly available in our GitHub repository.

\section{Results and Analysis}
\label{sec:results}

This section presents our empirical findings on multilingual dataset visibility. 
We first examine how catalogue-based RDI values are distributed across the 200 languages in our comparison set (Section~\ref{subsec:rdi-dist}). 
We then compare catalogue-based estimates with citation-based evidence of datasets appearing in the research literature (Section~\ref{subsec:catalog-vs-research}). 
Finally, we analyze properties of the discovered datasets, including their temporal emergence, accessibility, and task composition.

\label{subsec:rdi-dist}
\begin{figure}[t]
  \centering
  \includegraphics[width=\linewidth]{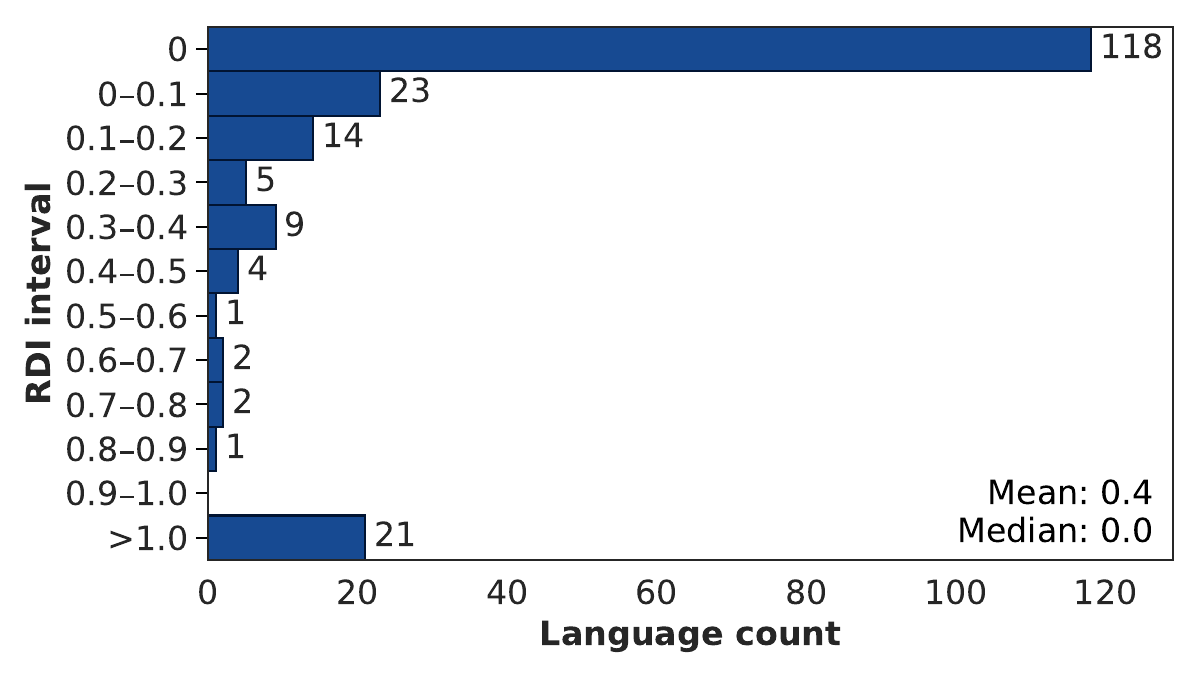}
  \caption{Distribution of average catalogue-based RDI across 200 high-population languages. The distribution is heavily skewed toward zero, with 118 languages (59\%) having no listed dataset in either catalogue, and a further 23 (11.5\%) fall below 0.1. Find details in the GitHub repository.}
  \label{fig:rdi-hist}
\end{figure}

\subsection{Distribution of Catalogue-based RDI}

Figure~\ref{fig:rdi-hist} shows the distribution of average catalogue-based RDI values for the 200 most widely spoken languages in our comparison set.
The distribution is highly skewed toward zero.
A total of 118 languages (59\%) have an average RDI of exactly 0.0, meaning that neither the LRE Map nor the LDC lists any dataset for them.
An additional 23 languages fall in the range $(0, 0.1)$.
Only 21 languages exceed an RDI of 1.0, corresponding to at least one catalogued dataset per million speakers.

The overall shape exhibits a sharp head–tail contrast.
A small subset of languages accounts for the majority of documented datasets, while the vast majority of high-population languages remain effectively absent from both catalogues.

This pattern motivates our focus on the sub-0.1 segment of the distribution, which represents the portion most likely to reflect documentation gaps rather than the genuine absence of dataset creation.

\subsection{Catalogue Coverage versus Research Usage}
\label{subsec:catalog-vs-research}

Our citation-based discovery and manual validation reveal substantial discrepancies between catalogue entries and datasets appearing in the research literature.
Across the 200 languages in our comparison set, catalogue records list only a few hundred distinct resources.
By contrast, our literature-based pipeline identifies 812 candidate dataset mentions across the 141 low-visibility languages examined in the citation analysis.
Of these candidates, 667 were verified as genuine dataset mentions, and 609 were retained as unique validated datasets after deduplication.
These 609 datasets span 53 languages, meaning that more than a quarter of the top-200 population languages show evidence of dataset activity in the research literature despite being absent or nearly invisible in the catalogue view.

\begin{table}[!tbp]
\centering
\resizebox{0.483\textwidth}{!}{%
\setlength{\tabcolsep}{4pt} 
\begin{tabular}{l|l|r|rr|rr|rr|r}
\toprule
\textbf{ISO}& & & \multicolumn{2}{c|}{\textbf{Our Method}} & \multicolumn{2}{c|}{\textbf{LRE Map}} & \multicolumn{2}{c|}{\textbf{LDC}} & \textbf{Avg.} \\
\textbf{639-3} & \textbf{Language} & \textbf{Pop.} & \textbf{Cnt.} & \textbf{RDI} & \textbf{Cnt.} & \textbf{RDI} & \textbf{Cnt.} & \textbf{RDI} & \textbf{RDI } \\
\midrule
\multicolumn{10}{l}{\textit{Pattern 1: Languages absent from catalogues but present in research}} \\
\midrule
tsn & Setswana & 13.7 & 26 & \textbf{1.90} & 0 & 0.00 & 0 & 0.00 & 0.00 \\
tat & Tatar & 4.8 & 7 & \textbf{1.46} & 0 & 0.00 & 0 & 0.00 & 0.00 \\
kir & Kyrgyz & 6.1 & 7 & \textbf{1.15} & 0 & 0.00 & 0 & 0.00 & 0.00 \\
npi & Nepali & 33.1 & 30 & \textbf{0.91} & 0 & 0.00 & 0 & 0.00 & 0.00 \\
bar & Bavarian & 13.7 & 12 & \textbf{0.88} & 0 & 0.00 & 0 & 0.00 & 0.00 \\
pst & Central Pashto & 7.3 & 6 & \textbf{0.82} & 0 & 0.00 & 0 & 0.00 & 0.00 \\
luo & Dholuo & 5.3 & 3 & \textbf{0.57} & 0 & 0.00 & 0 & 0.00 & 0.00 \\
sdh & Southern Kurdish & 6.0 & 3 & \textbf{0.50} & 0 & 0.00 & 0 & 0.00 & 0.00 \\
snd & Sindhi & 36.9 & 17 & \textbf{0.46} & 0 & 0.00 & 0 & 0.00 & 0.00 \\
ory & Odia & 39.5 & 16 & \textbf{0.41} & 0 & 0.00 & 0 & 0.00 & 0.00 \\
nso & Northern Sotho & 13.7 & 5 & \textbf{0.36} & 0 & 0.00 & 0 & 0.00 & 0.00 \\
pbu & Northern Pashto & 27.2 & 8 & \textbf{0.29} & 0 & 0.00 & 0 & 0.00 & 0.00 \\
tuk & Turkmen & 7.8 & 2 & \textbf{0.26} & 0 & 0.00 & 0 & 0.00 & 0.00 \\
mya & Burmese & 44.4 & 9 & \textbf{0.20} & 0 & 0.00 & 0 & 0.00 & 0.00 \\
dje & Zarma & 5.3 & 1 & \textbf{0.19} & 0 & 0.00 & 0 & 0.00 & 0.00 \\
tir & Tigrigna & 10.7 & 2 & \textbf{0.19} & 0 & 0.00 & 0 & 0.00 & 0.00 \\
syl & Sylheti & 11.5 & 2 & \textbf{0.17} & 0 & 0.00 & 0 & 0.00 & 0.00 \\
hae & Eastern Oromo & 12.1 & 2 & \textbf{0.17} & 0 & 0.00 & 0 & 0.00 & 0.00 \\
sot & Southern Sotho & 13.5 & 2 & \textbf{0.15} & 0 & 0.00 & 0 & 0.00 & 0.00 \\
pcm & Nigerian Pidgin & 120.7 & 16 & \textbf{0.13} & 0 & 0.00 & 0 & 0.00 & 0.00 \\
bho & Bhojpuri & 52.7 & 7 & \textbf{0.13} & 0 & 0.00 & 0 & 0.00 & 0.00 \\
nod & Northern Thai & 7.8 & 1 & \textbf{0.13} & 0 & 0.00 & 0 & 0.00 & 0.00 \\
xho & Xhosa & 19.2 & 2 & \textbf{0.10} & 0 & 0.00 & 0 & 0.00 & 0.00 \\
mag & Magahi & 21.0 & 2 & \textbf{0.10} & 0 & 0.00 & 0 & 0.00 & 0.00 \\
wes & Cameroon Pidgin & 12.0 & 1 & \textbf{0.08} & 0 & 0.00 & 0 & 0.00 & 0.00 \\
pan & Eastern Punjabi & 36.5 & 3 & \textbf{0.08} & 0 & 0.00 & 0 & 0.00 & 0.00 \\
sck & Sadri & 12.1 & 1 & \textbf{0.08} & 0 & 0.00 & 0 & 0.00 & 0.00 \\
run & Rundi & 12.9 & 1 & \textbf{0.08} & 0 & 0.00 & 0 & 0.00 & 0.00 \\
ctg & Chittagonian & 13.0 & 1 & \textbf{0.08} & 0 & 0.00 & 0 & 0.00 & 0.00 \\
tts & Northeastern Thai & 15.1 & 1 & \textbf{0.07} & 0 & 0.00 & 0 & 0.00 & 0.00 \\
nya & Chichewa & 14.5 & 1 & \textbf{0.07} & 0 & 0.00 & 0 & 0.00 & 0.00 \\
mai & Maithili & 17.6 & 1 & \textbf{0.06} & 0 & 0.00 & 0 & 0.00 & 0.00 \\
apd & Sudanese Arabic & 52.3 & 2 & \textbf{0.04} & 0 & 0.00 & 0 & 0.00 & 0.00 \\
lin & Lingala & 40.6 & 1 & \textbf{0.02} & 0 & 0.00 & 0 & 0.00 & 0.00 \\
pnb & Western Punjabi & 90.3 & 2 & \textbf{0.02} & 0 & 0.00 & 0 & 0.00 & 0.00 \\
\midrule
\multicolumn{10}{l}{\textit{Pattern 2: Languages significantly undercounted in catalogues}} \\
\midrule
ckb & Central Kurdish & 6.1 & 17 & \textbf{2.79} & 0 & 0.00 & 1 & 0.16 & 0.08 \\
asm & Assamese & 23.6 & 31 & \textbf{1.31} & 0 & 0.00 & 1 & 0.04 & 0.02 \\
ind & Indonesian & 252.4 & 196 & \textbf{0.78} & 31 & 0.12 & 3 & 0.01 & 0.07 \\
kin & Kinyarwanda & 15.3 & 12 & \textbf{0.78} & 0 & 0.00 & 1 & 0.07 & 0.03 \\
khm & Khmer & 19.0 & 14 & \textbf{0.74} & 0 & 0.00 & 3 & 0.16 & 0.08 \\
mar & Marathi & 99.3 & 41 & \textbf{0.41} & 14 & 0.14 & 0 & 0.00 & 0.07 \\
aka & Akan & 10.0 & 4 & \textbf{0.40} & 0 & 0.00 & 1 & 0.10 & 0.05 \\
uig & Uyghur & 13.6 & 5 & \textbf{0.37} & 0 & 0.00 & 1 & 0.07 & 0.04 \\
guj & Gujarati & 62.5 & 17 & \textbf{0.27} & 11 & 0.18 & 0 & 0.00 & 0.09 \\
som & Somali & 24.7 & 6 & \textbf{0.24} & 0 & 0.00 & 2 & 0.08 & 0.04 \\
swh & Swahili & 87.2 & 19 & \textbf{0.22} & 0 & 0.00 & 4 & 0.05 & 0.02 \\
yor & Yoruba & 49.9 & 10 & \textbf{0.20} & 0 & 0.00 & 2 & 0.04 & 0.02 \\
hau & Hausa & 94.4 & 16 & \textbf{0.17} & 12 & 0.13 & 2 & 0.02 & 0.07 \\
wol & Wolof & 17.3 & 3 & \textbf{0.17} & 0 & 0.00 & 1 & 0.06 & 0.03 \\
jav & Javanese & 69.2 & 11 & \textbf{0.16} & 0 & 0.00 & 1 & 0.01 & 0.01 \\
kmr & Northern Kurdish & 17.2 & 1 & 0.06 & 0 & 0.00 & 2 & 0.12 & \textbf{0.06} \\
ceb & Cebuano & 21.4 & 1 & 0.05 & 0 & 0.00 & 3 & 0.14 & \textbf{0.07} \\
nan & Min Nan Chinese & 45.8 & 2 & \textbf{0.04} & 0 & 0.00 & 7 & 0.15 & \textbf{0.08} \\
\bottomrule
\end{tabular}
}
\caption{Comparison between citation-mined dataset evidence and catalogue records from the LRE Map and LDC. 
Cnt. denotes the number of datasets, Pop. the speaker population (in millions), and RDI the number of datasets per million speakers. 
Pattern~1 shows languages absent from both catalogues but with datasets identified in the literature, while Pattern~2 highlights languages whose dataset activity in research substantially exceeds catalogue counts. 
Avg. RDI is the mean of the LRE Map and LDC estimates.}
\label{tab:rdi-comparison}
\end{table}

Table~\ref{tab:rdi-comparison} provides a detailed comparison of RDI values between catalogue records and our citation-based dataset inventory.
Two recurrent patterns emerge.

\paragraph{Languages absent from catalogues but present in the Research literature.}
We identify 35 languages that have an average RDI of 0.0 in both the LRE Map and the LDC, yet appear with non-zero RDIs in our citation-based evidence.
Examples include Setswana (26 datasets; RDI = 1.90), Tatar (7 datasets; RDI = 1.46), Nepali (30 datasets; RDI = 0.91), and Sindhi (17 datasets; RDI = 0.46).
These languages transition from catalogue absence to documented presence once literature-based evidence is considered.

\paragraph{Severe undercounting in catalogues.}  
For many languages that do have catalogue entries, the recorded counts are substantially lower than the numbers suggested by the research literature.
For example, Indonesian rises from 34 listed resources across catalogues (average RDI = 0.07) to 196 research datasets (RDI = 0.78);
Marathi from 14 listed (RDI = 0.07) to 41 research datasets (RDI = 0.41);
and Assamese from 1 listed (RDI = 0.02) to 31 datasets (RDI = 1.31).
These differences indicate that catalogue-based estimates can substantially underestimate the number of datasets visible in the research literature for many high-population languages.

The increase in observed datasets through citation-based discovery suggests that multilingual resource inequality cannot be understood solely in terms of whether datasets exist, but also in terms of whether they are visible and discoverable through commonly used documentation channels.
In several cases, resources produced and reused within regional research communities, for example, in Indonesia, India, and parts of southern Africa, remain outside major catalogue registries despite appearing in multiple research papers.

\label{subsec:temporal-trends}
\begin{figure}[t]
  \centering
  \includegraphics[width=1\linewidth]{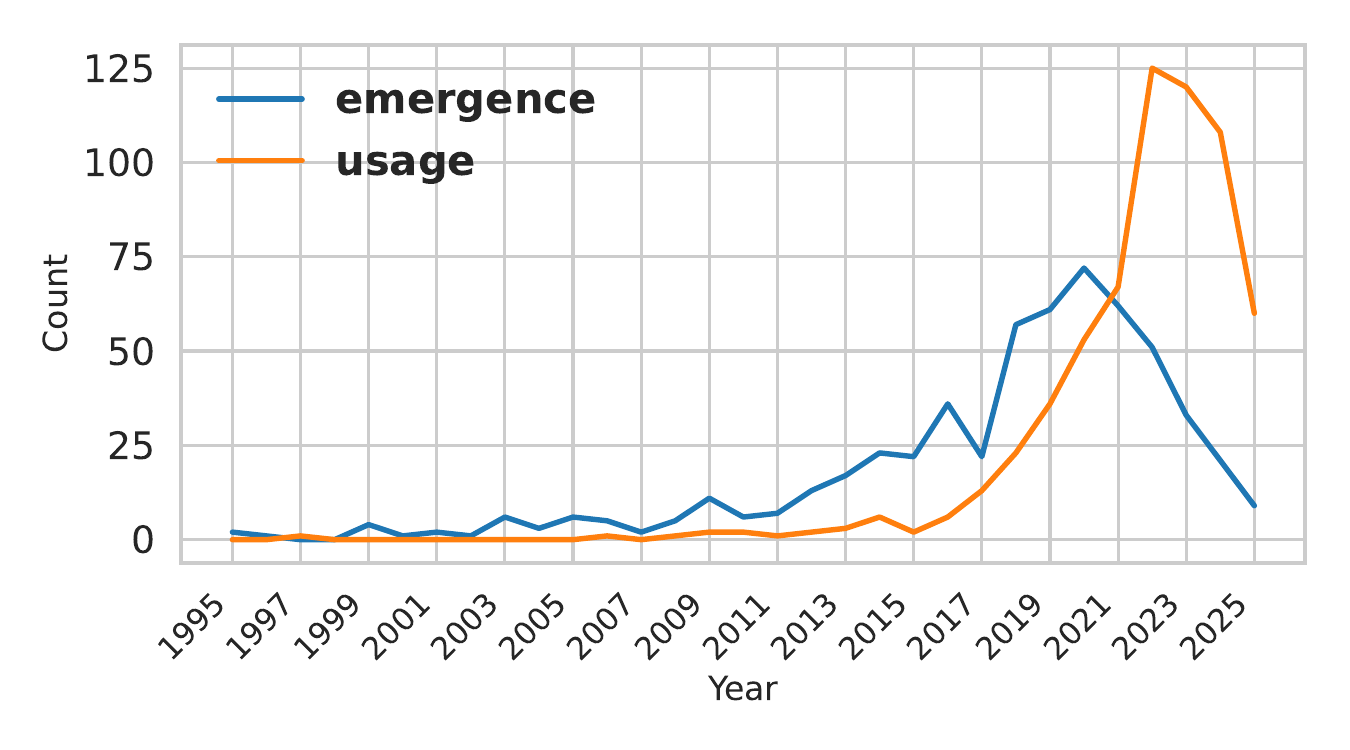}
  \caption{Aggregate trends in dataset emergence (release year) and subsequent usage (citation year). Use typically trails emergence by one to two years.}
  \label{fig:agg-appearance-usage}
\end{figure}

\subsection{Temporal and Usage Trends}

For each dataset, we record both its emergence year (the publication year of the canonical source paper, if available) and its usage years (the publication years of papers citing or reusing it).
Figure~\ref{fig:agg-appearance-usage} aggregates these counts across all languages.
The global trend shows a consistent emergence–usage lag of roughly one to two years, indicating that newly released datasets typically begin to appear in subsequent research within a short window.
The emergence curves rise steadily over the last decade, with a visible peak around 2020, followed by a modest apparent decline through 2025.
The usage curves show a similar pattern with a peak around 2022–2023.
This apparent decline may partly reflect citation latency, which can delay the observable uptake of recently released datasets.

\begin{figure*}[t]
    \centering
    \includegraphics[width=0.95\linewidth]{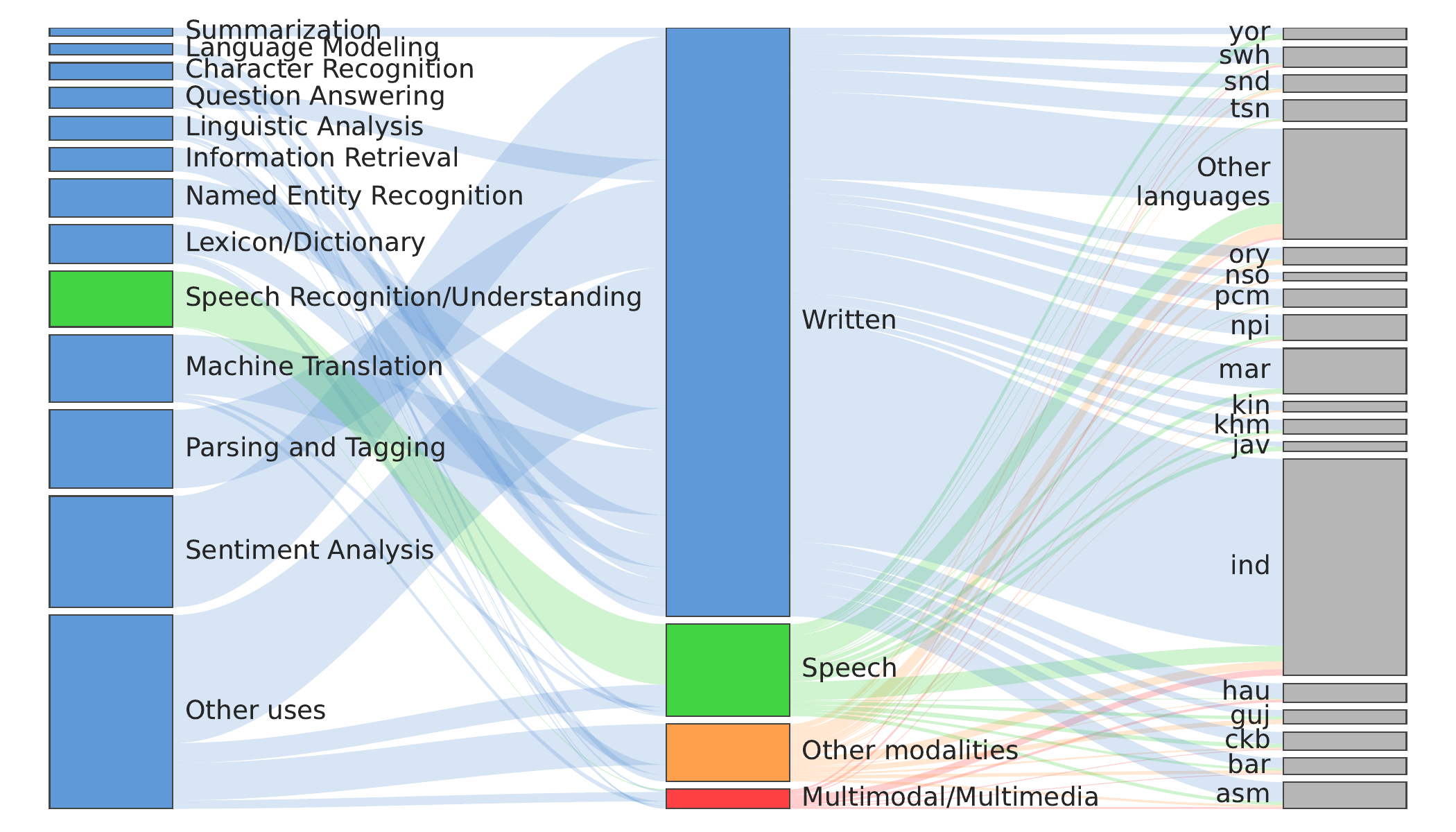}
    \caption{Summary of tasks, modalities, and languages for the 609 discovered datasets, showing the flow from NLP task to data modality and target language. The width of the flows is proportional to the number of datasets.}
    \label{fig:modality}
\end{figure*}

\subsection{Dataset Distribution by Task, Modality, and Language}
\label{subsec:modality}

Figure~\ref{fig:modality} visualizes the distribution of the 609 discovered datasets, mapping their intended NLP tasks (left) to their data modalities (center) and target languages (right).
Most datasets are text-based.
Foundational tasks such as Sentiment Analysis, Parsing and Tagging, and Machine Translation are overwhelmingly associated with written corpora.
This pattern is particularly visible for languages such as Indonesian (ind), Yoruba (yor), Swahili (swh), and Sindhi (snd).
Speech-based resources are largely concentrated in tasks related to speech recognition and spoken language understanding, and are rarely reused for other purposes.
Multimodal datasets are comparatively rare.

The distribution also shows differences in modality emphasis across languages.
For example, languages such as Hausa (hau), Assamese (asm), Gujarati (guj), and Central Kurdish (ckb) have relatively stronger representation in speech-oriented datasets, while other languages appear primarily in text-based resources.
These differences suggest that dataset creation efforts may be shaped by distinct research communities or regional initiatives that emphasize particular modalities.

\subsection{Aggregate Patterns and Interpretation}
\label{subsec:aggregate-patterns}
Taken together, these findings highlight two distinct aspects of multilingual dataset inequality:
(i) catalogue-based metrics can miss datasets that exist and circulate in the research literature but have never been registered in major repositories. 
This documentation gap means that catalogue statistics alone may underrepresent the presence of datasets for some languages.
(ii) even when datasets are documented, their practical accessibility varies widely. 
Some datasets remain openly downloadable, while others become difficult to obtain due to dead links, restricted distribution, or institutional licensing conditions. 
As a result, the existence of a dataset in the literature or in a catalogue does not necessarily imply that it is readily usable by new researchers.

Languages such as Indonesian, Marathi, and Assamese exhibit substantially higher dataset counts in the research literature than their catalogue records suggest, while others such as Finnish remain both well documented in catalogues and frequently used in research.
This asymmetry suggests that the low-resource label, when assigned solely from catalogue data, can conflate true scarcity with documentation and visibility effects.
Our results therefore complement prior qualitative audits such as \citet{yu-etal-2022-beyond} by providing a quantitative, usage-centered perspective on how multilingual datasets are created, recorded, and circulated in practice.

\section{Conclusion}

This study revisits how multilingual NLP conceptualizes low-resource status.
Rather than equating scarcity with the absence of data, we examine how the visibility of datasets is shaped by documentation practices and research circulation.

To do so, we combine two complementary perspectives.
First, we introduce the Resource Density Index (RDI), a population-normalized measure of catalogue visibility across the 200 most widely spoken languages.
Second, we construct a citation-based inventory of datasets appearing in the research literature and manually validate their existence and accessibility.
Comparing these two perspectives reveals systematic discrepancies between catalogue documentation and research practice.
Many languages with large speaker populations appear nearly invisible in major catalogues, yet show clear evidence of dataset activity in the literature.
These findings suggest that part of what is sometimes interpreted as data scarcity may instead reflect gaps in documentation, indexing, and discoverability.

At the same time, our analysis shows that dataset existence does not necessarily imply dataset accessibility.
Even when datasets can be identified through publications, a substantial portion cannot be readily obtained due to dead links, restricted distribution, or other access barriers.
Multilingual dataset inequality therefore involves not only the creation of resources, but also the conditions under which those resources remain visible and usable over time.
By integrating demographic normalization with literature-based evidence, our framework provides a practical way to surface under-documented datasets and improve their discoverability.
Recognizing the distinction between scarcity and visibility may help the field refine how “low-resource” is defined, how dataset coverage is measured, and how funding and benchmark attention are distributed in multilingual NLP.

\section{Future Work}

Future work will extend this framework beyond the low-visibility segment examined here to languages with higher catalogue RDIs. 
A preliminary run over the full set of 200 languages retrieves 7,299 candidate dataset mentions prior to manual validation, indicating that a substantial portion of the multilingual dataset landscape remains to be systematically verified and documented.

A second direction is to move beyond the predominantly English-language research literature. 
Incorporating non-English academic corpora and regional repositories would allow dataset discovery closer to the sites of resource production, yielding a more faithful account of the global data ecosystem while enabling broader discovery, sharing and circulation of multilingual resources.

\nocite{*}

\section{Limitations}
While our study reveals substantial asymmetries in multilingual dataset visibility, we acknowledge several limitations in our methodology and scope that offer avenues for future research.
Our dataset discovery process, while effective, has inherent constraints that may lead to an underestimation of the true resource landscape.

Our retrieval stage also operates under a bounded search window. 
For each language, we collect citation contexts from at most the top $k=400$ relevant papers in Semantic Scholar before expanding through their references and citations. 
This limit keeps the discovery process computationally tractable and focuses the search on the most visible portion of the literature. 
However, it inevitably constrains recall. 
For languages with very large research footprints, such as English, even substantially larger thresholds (e.g., $k=1000$) would still cover only a small fraction of all potentially relevant papers. 
Increasing this window would likely surface additional dataset mentions for some languages, suggesting that our inventory should be interpreted as a conservative lower bound rather than an exhaustive catalogue of multilingual datasets.

The pipeline's effectiveness is contingent on the explicitness of academic writing. It primarily identifies dataset-language associations when the language is explicitly named in the citation context. For instance, if a paper utilizes a multilingual dataset containing five languages but only mentions three in its description, our pipeline would fail to associate the dataset with the other two languages. We observed this in practice, where a single dataset was linked to fewer languages than it actually contained, leading to an underestimation of its multilingual reach.

While we performed manual normalization of language names to align with \textit{Ethnologue}, this process is not exhaustive. A single language may be referred to by multiple names, such as endonyms, exonyms, dialectal variants, or historical names. Our normalization may not capture all such variations, potentially leading to fragmented or incomplete counts for some languages.

Our analysis is predominantly based on the English-language portion of the Semantic Scholar corpus. We are therefore likely to miss datasets that are described exclusively in papers written in other languages. During our validation, for example, we identified several key Indonesian datasets that were first introduced and discussed in papers published in Indonesian. This highlights a systematic bias in our discovery process that favors anglophone research communities and underreports resources from other linguistic spheres.

Furthermore, our characterization of dataset usage is based on how resources are applied in the literature, not necessarily on their original intended purpose. The task labels for our discovered datasets (visualized in Figure~\ref{fig:modality}) are automatically extracted from the context of the citing papers. Consequently, our analysis captures the actual application of a dataset, which can differ from its original design. For example, a corpus created for syntactic parsing might be repurposed for a sentiment analysis task. Our method would categorize it based on the latter usage, reflecting its evolving role in the community but potentially obscuring its original function.

Our analysis also identifies a significant number of datasets that are documented but not openly available. It does not, however, extend to the root causes of this inaccessibility. The reasons are likely multifaceted, ranging from restrictive institutional licensing, privacy concerns over the data subjects, to the natural decay of URLs and hosting infrastructure over time. While we recorded whether datasets are openly downloadable, a finer-grained analysis of license types (e.g., non-commercial restrictions, request-for-access agreements) and their implications for practical reuse remains an important direction for future work. Similarly, our literature-centric approach systematically underrepresents resources from academic grey literature, such as datasets developed for Master's or PhD theses, which often lack formal publication and hosting mechanisms.

Our catalogue baseline is drawn from the LRE Map and the LDC, two long-standing and widely cited resource catalogues. We do not include more recent platforms such as the Hugging Face Datasets Hub or OpenSLR. Hugging Face, while increasingly popular, operates as an open-upload platform where community-contributed entries lack the editorial curation and metadata standardization of institutional catalogues, making systematic comparison difficult without additional verification of each entry's provenance, language coverage, and quality. OpenSLR, though valuable, is specialized in speech resources for a single modality and does not aim to provide comprehensive multilingual coverage across task types. Extending the baseline to incorporate these and other emerging platforms, after appropriate quality filtering, is a natural next step.


\section{Acknowledgment}
Zhiyin Tan was funded by the ``HybrInt - Hybrid Intelligence through Interpretable AI in Machine Perception and Interaction'' project (Zukunft Nds, Niedersächsisches Ministerium für Wissenschaft, Grant ID: ZN4219).

Changxu Duan was funded by the InsightsNet project (funded by the Federal Ministry of Education and Research (BMBF) under grant no. 01UG2130A). 

We gratefully acknowledge support from the hessian.AI Service Center (funded by the Federal Ministry of Research, Technology and Space, BMFTR, grant no. 16IS22091) and the hessian.AI Innovation Lab (funded by the Hessian Ministry for Digital Strategy and Innovation, grant no. S-DIW04/0013/003).

We gratefully acknowledge the Research group Corpus and Computational Linguistics, Institute of Linguistics and Literary Studies, TU Darmstadt.

We gratefully acknowledge the computing time granted by the Resource Allocation Board and provided on the supercomputer Emmy/Grete at NHR-Nord@Göttingen as part of the NHR infrastructure. The calculations for this research were conducted with computing resources under the project \verb|nhr_he_starter_25563|.

\section{Bibliographical References}\label{sec:reference}

\bibliographystyle{lrec2026-natbib}
\bibliography{ref}
\bibliographylanguageresource{languageresource}

\end{document}